\documentclass[sigconf]{acmart}
\usepackage[table]{xcolor}
\AtBeginDocument{%
  }

\copyrightyear{2026}
\acmYear{2026}
\setcopyright{cc}
\setcctype{by}
\acmConference[SIGIR '26]{Proceedings of the 49th International ACM SIGIR Conference on Research and Development in Information Retrieval}{July 20--24, 2026}{Melbourne, VIC, Australia}
\acmBooktitle{Proceedings of the 49th International ACM SIGIR Conference on Research and Development in Information Retrieval (SIGIR '26), July 20--24, 2026, Melbourne, VIC, Australia}
\acmDOI{10.1145/3805712.3808458}
\acmISBN{979-8-4007-2599-9/2026/07}

\begin{document}

\title{A Cascaded Generative Approach for e-Commerce Recommendations}

\author{Moein Hasani}
\authornote{Both authors contributed equally to this work.}
\affiliation{%
  \institution{Instacart}
  \city{Toronto}
  \country{Canada}}
\email{moein.hasani@instacart.com}
\orcid{0009-0009-4117-4289}

\author{Hamidreza Shahidi}
\authornotemark[1]
\affiliation{%
  \institution{Instacart}
  \city{Toronto}
  \country{Canada}}
\email{hamidreza.shahidi@instacart.com}
\orcid{0009-0007-0189-5597}

\author{Trace Levinson}
\affiliation{%
  \institution{Instacart}
  \city{Brooklyn}
  \country{USA}}
\email{trace.levinson@instacart.com}
\orcid{0009-0004-6704-4363}

\author{Yuan Zhong}
\affiliation{%
  \institution{The Pennsylvania State University}
  \city{University Park}
  \country{USA}
}
\email{yfz5556@psu.edu}
\orcid{0009-0009-4427-5667}

\author{Guanghua Shu}
\affiliation{%
  \institution{EvenUp}
  \city{San Francisco}
  \country{USA}}
\email{guanghua.shu@evenup.ai}
\orcid{0000-0001-7972-8616}

\author{Vinesh Gudla}
\affiliation{%
  \institution{Ambience Healthcare}
  \city{San Francisco}
  \country{USA}}
\email{vinesh.gudla@ambiencehealthcare.com}
\orcid{0009-0000-3183-1921}

\author{Tejaswi Tenneti}
\affiliation{%
  \institution{Ambience Healthcare}
  \city{San Francisco}
  \country{USA}}
\email{tejaswi.tenneti@ambiencehealthcare.com}
\orcid{0009-0008-7301-6958}

\renewcommand{\shortauthors}{Moein Hasani et al.}

\begin{abstract}
Personalized storefronts in large e-commerce marketplaces are often assembled from many independent components: static themes per page section ("placement"), retrieval systems to fetch eligible products per placement, and pointwise rankers to order content. While effective in optimizing for aggregate preferences, this paradigm is rigid and can limit personalization and semantic cohesion across the page. This makes it poorly suited to support dynamic objectives and merchandising requirements over time.

To address this, we introduce a cascaded merchandising framework that decomposes storefront construction into two generative tasks: (i) placement-level theme generation and (ii) constrained keyword generation per placement to power product retrieval. Teacher–student fine-tuning is leveraged to improve scalability of this framework under production latency and cost constraints. Fine-tuned model ablations are shown to approach closed-weight LLM performance. We further contribute frameworks for AI-driven content evaluation and quality filtering, enabling safe and automated deployment of dynamic content at scale. Generative output is fused with traditional ranking models to preserve hybrid infrastructure. In online experiments, this framework yields an estimated +2.7\% lift in cart adds per page view over a strong baseline.
\end{abstract}




\maketitle

\section{Introduction}
Storefronts are a central surface in e-commerce, acting as the main hub for product discovery and carrying strong impact on user engagement and long-term retention. In many production systems, storefronts are built via a modular pipeline:
\begin{enumerate}
\item Human merchandisers predefine content, such as a "Dairy" header, and product retrieval sources per page section (each section will hereafter be referred to as a \textbf{placement}, often a carousel of products).
\item Retrieval systems fetch products per placement.
\item Ranking models order placements and items to optimize a fixed set of business metrics.
\end{enumerate}

This pattern can perform well in optimizing for aggregate preferences. However, a one-size-fits-all content library eventually reaches diminishing returns. Fixed semantic structure and a lack of cross-placement awareness make it difficult to deepen personalization or target page-level objectives such as coherence and diversity.

To solve for these limitations, we propose a multi-phased generative merchandising framework to redefine storefront construction, targeting a balance between personalization and production stability. Remaining sections will walk through our contributions in detail. We first introduce the problem statement and background more formally. Methodology and architecture are outlined with detail on major components, including Retrieval-Augmented Generation (RAG)-based retrieval \cite{lewis2020rag, gao2024rag}, teacher-student fine-tuning, and in-house evaluators. Finally, we summarize key offline and online results, with online experiments deployed to production users within a large grocery e-commerce marketplace.

\section{Background}
\subsection{Limitations of Traditional Storefront Recommendation Engines}
As introduced above, traditional recommender architectures suffer from key limitations:

\paragraph{Difficulty in scaling personalized content.} New content is generally conceived manually for a specific objective. Once created, each placement becomes universally eligible to all users for serving. This human-driven process is expensive and time intensive, with generation and evaluation of content managed by hand. As a result, traditional architectures inhibit the ability to deeply personalize the storefront – not only per user, but also according to seasonal and other transient dimensions.
\paragraph{Lack of cohesion.} As multiple merchandisers generate content in silos, the series of placements displayed to a user can result in a chaotic surface presentation. Users are required to scroll without the ability to easily navigate the page to solve their needs.

\subsection{Introducing Generative Methods for Recommendation
}
Large language models offer a natural mechanism for producing cohesive, dynamic, and personalized content. As such, we introduce generative models to solve for the  challenges above. Two high-level paradigms are considered:
\begin{enumerate}
\item \textit{Bottoms-up generation:} Directly generate all possible products to serve to a user through sequential token prediction, then cluster and organize them across the storefront. 
\item \textit{Top-down generation:} First generate ordered placement themes to structure the page, then generate entities to power product retrieval per placement.
\end{enumerate}

Bottoms-up generation contains the benefit of deep flexibility with less constrained recall. However, it can be difficult to deploy at scale: large vocabularies, frequent catalog changes, and latency and cost constraints complicate inference. Further, with a much broader modeling task, it can be difficult to ensure generated entities meet a diverse set of page requirements, and may require significant fine-tuning efforts as needs evolve. 

For the storefront use case, we find these realities favor the top-down, cascaded architecture introduced in Section 4.

\subsection{Related Work}
The broader landscape of generative recommendation, spanning LLM-based, diffusion, and sequential approaches, is surveyed in \cite{hou2025survey}. Recent work has explored a number of generative approaches for slate and page-level recommendations. Tomasi et al. \cite{tomasi2025} introduced diffusion models for prompt-conditioned slate generation in music streaming, demonstrating that generative models can produce coherent item sets from natural language descriptions. While their approach generates slates from prompts, our work adopts an end-to-end strategy to design placement themes across the full page and then produce retrieval keywords—enabling adaptability to a large suite of business requirements.

TIGER was one of the foundational works demonstrating how generative models can be applied to recommendation by reframing retrieval as sequence generation: items are represented via semantic IDs and a transformer autoregressively decodes these identifiers from user interaction context \cite{tiger2023}. This corresponds to the bottoms-up approach described in Section 2.2 above, rather than our top-down approach at the semantic planning level. At industrial scale, Zhai et al. \cite{zhai2024hstu} propose HSTU, a trillion-parameter sequential transducer that reformulates recommendation as generative sequence transduction over user actions. While demonstrating strong results, this bottoms-up approach requires massive scale to handle high-cardinality item vocabularies — further motivating our top-down cascaded design which avoids direct item generation.

Maragheh et al. \cite{arag2025} introduced ARAG, an agentic RAG framework using specialized agents for user understanding and semantic alignment in personalized recommendations. Our RAG approach differs by using embedding-based retrieval to constrain keyword generation within a curated taxonomy, balancing recall with relevance. Sun et al. \cite{Sun2024APQ} proposed Product-RAG, a RAG framework for e-commerce query auto-completion that retrieves catalog products given a search prefix and conditions a generative LLM on the retrieved metadata to produce product-aware suggestions. Like our Phase 2 approach, Product-RAG grounds LLM generation in catalog-aware retrieval to reduce hallucinations; however, their work targets query suggestion from partial prefixes, whereas ours applies RAG to constrain keyword generation for personalized placement themes.

For evaluation, Fabbri et al. \cite{fabbri2025} proposed profile-aware LLM-as-a-judge \cite{mehrdad2024llm, zheng2023judging} methods for podcast recommendations, constructing natural language user profiles to assess recommendation quality at scale. Similarly, we employ LLM-based evaluators across multiple levels (page, placement, and product) but supplement them with fine-tuned cross-encoders for production-scale filtering.

\section{Problem Formulation}
Let $c_u$ be the context for user u (purchase history, engagement signals, dietary and other derived household preferences). The goal is to generate a user-personalized storefront $H_u, \forall u \in U$ for eligible user universe $U$, composed of an ordered array of $m$ placements:
\begin{equation}
H_{u} = [z_{iu}(t_{iu}, s_{iu})]_{i=1}^{m},
\label{eq:storefront}
\end{equation}
where each placement $z_{iu}$ consists of a semantic intent theme $t_{iu}$ along with a corresponding slate of products $s_{iu}(p\in P)$ for product universe $P$.

Our goal is to personalize both page structure and content per placement. We aim to maximize page-level user utility:
\begin{equation}
Y_u(H_u \mid c_u, P, B)
\label{eq:utility}
\end{equation}
where $B$ is a set of latency, business, and policy constraints (ex: specific growth objectives or brand compliance rules). $Y$ is defined by a multi-component objective to balance short-term and long-term user engagement.

\section{Model Architecture}
\label{sec:architecture}

\subsection{Overview}
The recommendation pipeline consists of four sequential phases:
\begin{enumerate}
  \item Page design and theme generation
  \item Retrieval keyword generation
  \item Quality and diversity filtering
  \item Item and pagewise ranking
\end{enumerate}
The first three phases comprise our generative content architecture, shown in Figure 1. Phase 4 invokes the output of this pipeline for serving integration with existing ranking systems.

\begin{figure}[t]
  \centering
  \includegraphics[width=\linewidth]{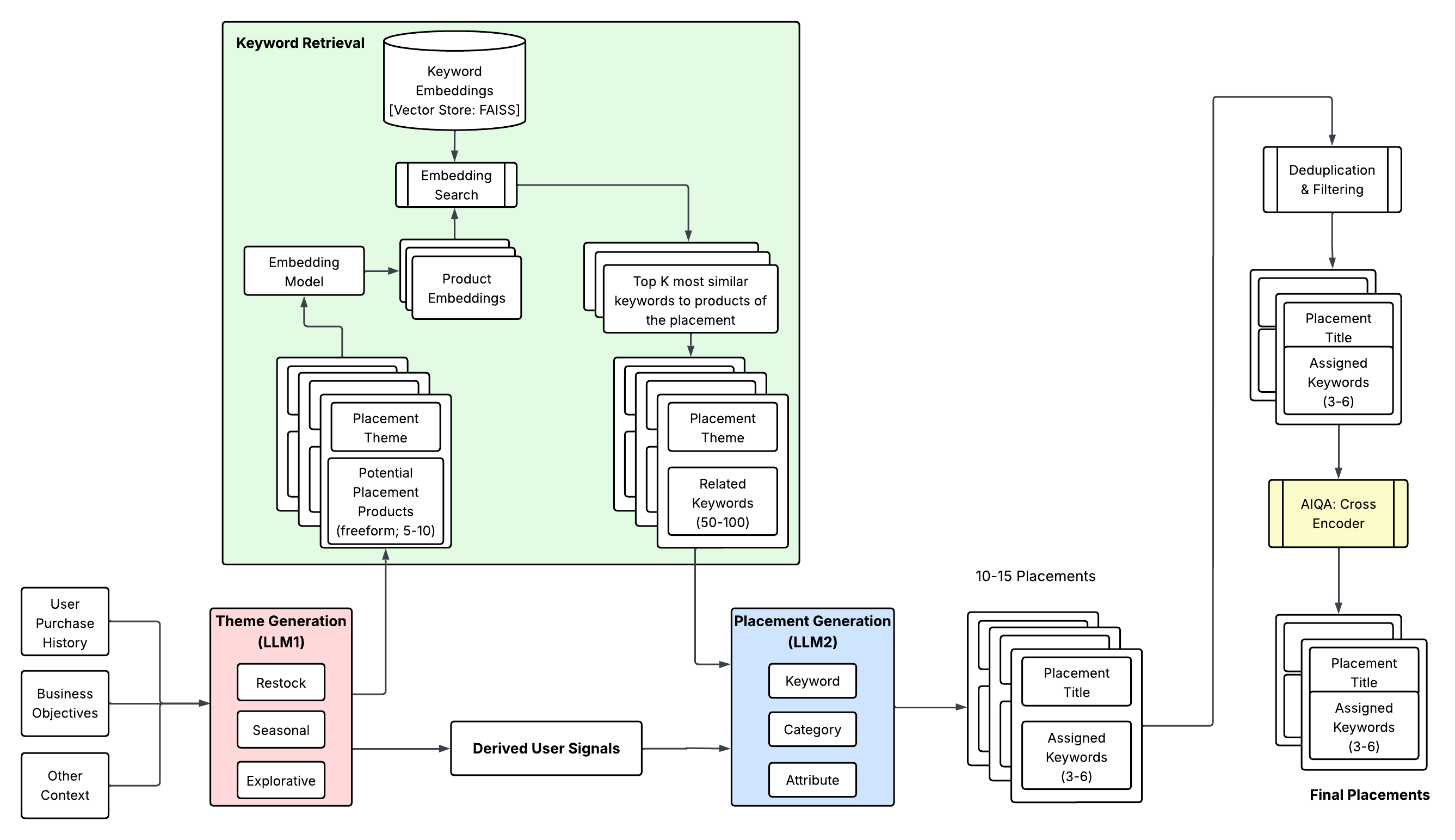}
  \caption{Cascaded generative content architecture. LLM1 generates personalized placement themes from raw signals. LLM2 combines these themes with RAG‑retrieved candidate keywords to generate final keywords per placement. Output undergoes post-processing and AI Quality Assurance (AIQA) filtering for quality, diversity, and policy compliance.
}
  \label{fig:architecture}
\end{figure}

\subsection{Page Design and Theme Generation}
First, a page design agent leverages user context $c_u$  to produce a set of $m$ high-level themes personalized to each user $\{t_{iu}\}_{i=1}^{m}$ that seek to maximize $Y_u$. Each theme represents a discrete and coherent shopping intent, which in turn serves as a placement title (ex: "Flavor builders for weeknight meals" or "Functional hydration, lower sugar"). We implement theme generation using a closed-weight LLM, leveraging constrained decoding \cite{willard2023constrained} and structured output to ensure downstream usability. Future work will explore fine-tuning under the methodology of Section 4.3.1.

To optimize downstream token efficiency, LLM1 outputs both placement entities and derived signals, such as user personas and freeform product concepts that align well with placement intent. This removes the need to ingest redundant raw signals in Phase 2.

Guardrails are then placed against LLM1 output to enforce business and policy constraints. Deterministic fallback behavior takes over when generation fails or produces low-quality output.

\subsection{Retrieval Keyword Generation}
Once generated in Phase 1, each theme $t_{iu}$ is fed as context to generate one or more retrieval-compatible descriptors (ex: search query strings; catalog taxonomy nodes) used to retrieve slate $s_{iu}$. Multiple descriptor representations are explored, with structured nodes proving most performant. For simplicity, we will refer broadly to these descriptors as \textbf{keywords}.

\subsubsection{Teacher-Student Fine-Tuning}
To optimize latency and cost, we leverage teacher–student distillation \cite{hinton2015distilling}: a closed-weight LLM generates high-quality supervised data, human-validated on a small sample. An internal model is then fine-tuned with LoRA adapters \cite{hu2021lora} to imitate the teacher's behavior. 

Several ablation studies are performed to converge toward the optimal student model:
\begin{itemize}
\item Multiple base model explorations within each of the Llama and Qwen \cite{qwen2025} families
\item LoRA adapter rank variation
\item Fine-tuning sample size augmentation
\item Label quality filtering with AIQA
\end{itemize} 

Strongest results are achieved using a Llama‑3.2‑3B \cite{grattafiori2024llama3} student model, with performance approaching parity with the teacher model. Comparative results across a subset of these ablations are summarized in Table 1.

\subsubsection{RAG}
To further improve prompt efficiency while maintaining precision, we incorporate RAG into the Phase 2 pipeline. First, embeddings are generated for the freeform product concepts produced during LLM1 inference. Eligible candidate keywords per theme are then restricted using embedding‑based similarity. Nearest neighbors are retrieved from a 300,000‑term keyword corpus, and only this refined subset is fed as candidates to LLM2 for constrained generation. The introduction of RAG reduces total inference cost by 15-20\% in production settings. This is a key motivator for our cascaded architecture, as a single LLM would require passthrough of the full keyword corpus as context.

\subsection{Quality and Diversity Filtering}
Given the dynamic nature of this system, additional guardrails are implemented to uphold diversity and relevance of final placements. The system handles this in two stages:

\paragraph{Semantic deduplication.}
Embeddings are generated for each resulting theme, and similarity-thresholded deduplication is applied to remove redundant themes per user. 

\paragraph{Fine-tuned cross encoder for product-theme relevance.}
A cross encoder (DeBERTa-v3-base architecture \cite{he2021debertav3}) is fine-tuned to enforce relevance at scale. This model is trained on a set of ground truth pairs for theme-product relevance, generated via a closed-weight teacher model with human-in-the-loop validation. We observe strong student performance at over 90\% alignment with ground truth labels. At runtime, any generated placements that this model identifies as quality violations are pruned before deployment. Relative to its closed-weight LLM-as-a-judge equivalent, this model reduces inference cost by more than 99\%, enabling filtering across all users within online experiments.

\subsection{Item and Pagewise Ranking}
Once the list of ordered placements and keywords is finalized, output is cached for low-latency retrieval at runtime. Existing product and placement ranking services retrieve generated content, perform any additional re-ranking, and return final placements on the page. This design modularizes the ecosystem. Generative retrieval is decoupled from traditional ranking services, while maintaining a path to deeper generative control as the system matures.

\section{Product Experience}
This merchandising framework meaningfully shifts placement composition. Illustrative examples of legacy vs. generative placements are shown in Figures 2 and 3. Generative merchandising replaces rigid content with engaging thematic placements, each retrieving several underlying product categories personalized to the user (e.g. meats, cheeses, and breads).

\begin{figure} 
  \centering
  \includegraphics[width=\linewidth]{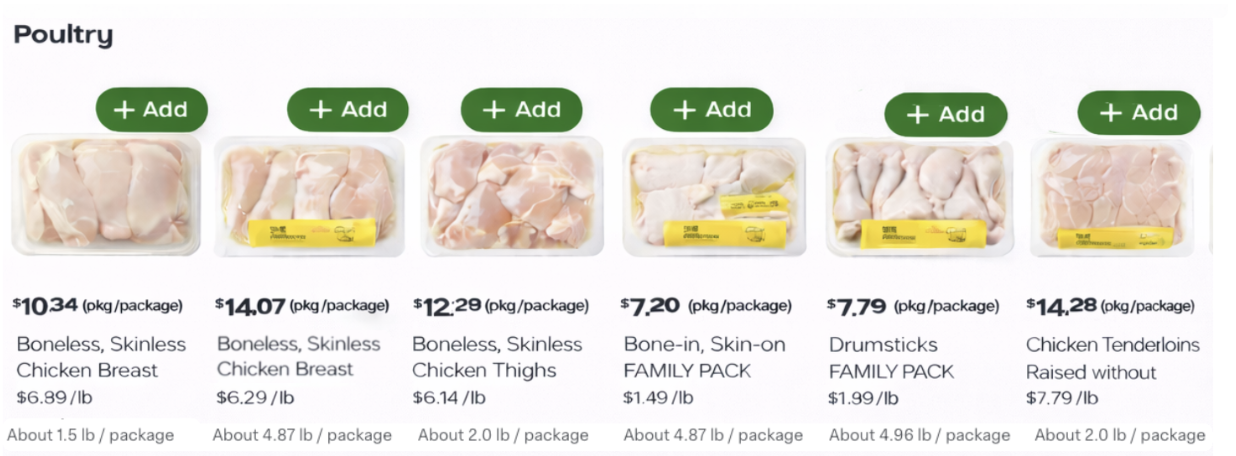}
  \caption{Example control experience (illustrative)}
  \label{fig:architecture}
\end{figure}

\begin{figure} 
  \centering
  \includegraphics[width=\linewidth]{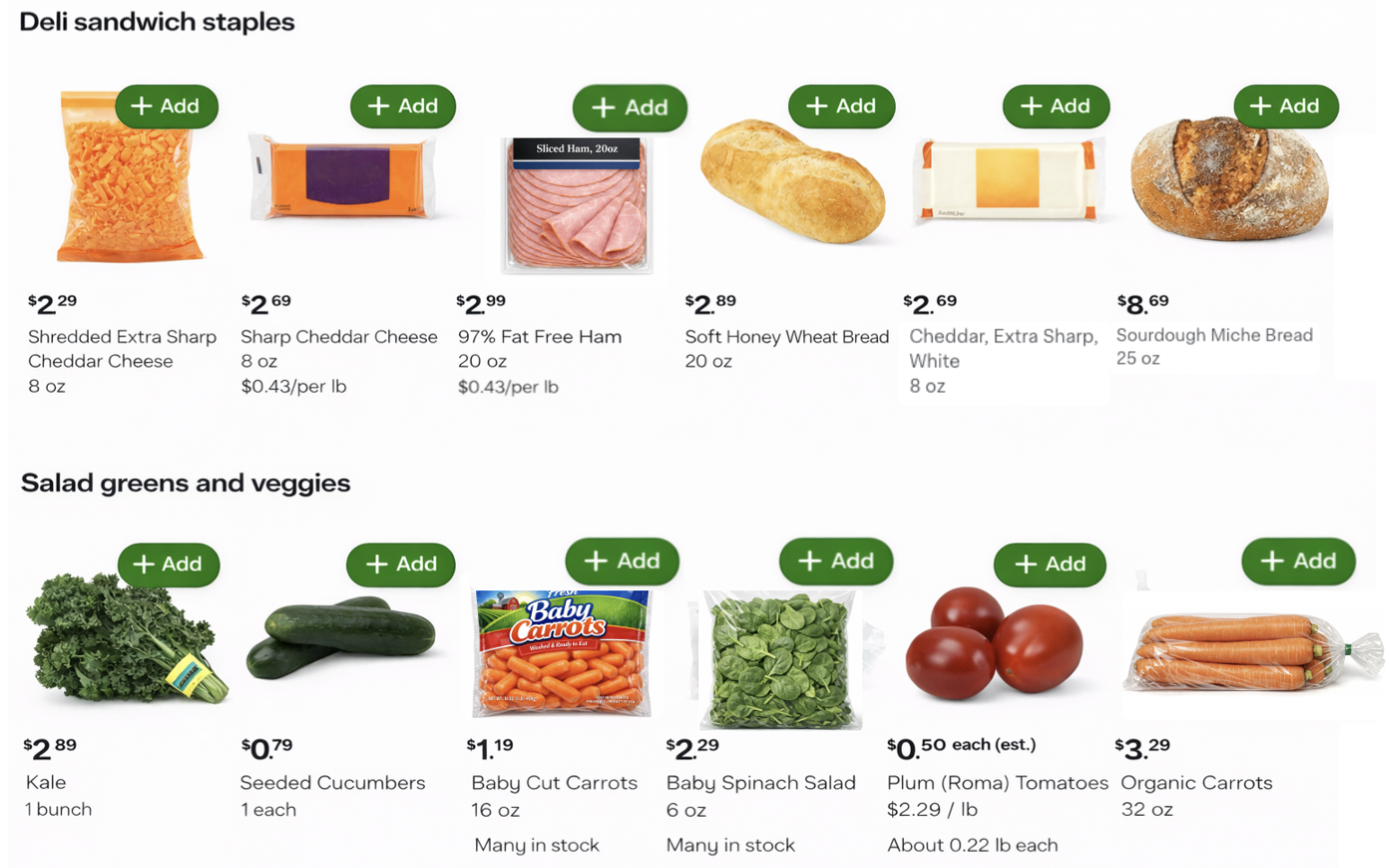}
  \caption{Example treatment experience (illustrative)}
  \label{fig:architecture}
\end{figure}

\section{Experiments and Results}

We introduce a hybrid offline--online evaluation framework to compare the performance of generative recommendations against the production baseline. The dataset and product catalog used in this work are proprietary, which limits direct reproducibility of our experimental results. However, the cascaded generative architecture, teacher--student fine-tuning pipeline, RAG-based retrieval methodology, and multi-level AIQA evaluation framework are all designed to be transferable to other e-commerce domains.

\subsection{Offline Evaluation}
We develop offline evaluators to assess content quality across multiple dimensions, with key results summarized in Table~\ref{tab:offline_results}.

\paragraph{LLM-as-a-Judge Evaluators.}
We develop LLM-as-a-judge evaluators to assess quality along each level of the content hierarchy: (1)~page-level, (2)~placement-level, and (3)~product-level. These evaluators are deployed against a random sample of thousands of users, held consistent across all policies to ensure robust cross-policy comparisons.

\paragraph{Metric-Based Evaluations.}
We incorporate a number of metric-based signals to supplement the LLM judges, including recall density and existing machine learning scores such as $p(\text{Conversion})$.

\begin{table}
\centering
\caption{Offline evaluation results. P-T@K: product-theme precision (Average \% of top-K products relevant to placement title across placements). K-U: keyword-user precision (\% keywords relevant to user profile across all placements). Density: average product recall volume per placement. All student models undergo supervised fine-tuning.}
\label{tab:offline_results}
\begin{tabular}{lcccc}
\textbf{Model} &
\textbf{P-T@5} &
\textbf{P-T@20} &
\textbf{K-U} &
\textbf{Density} \\
\hline
Control (production baseline) &
0.824 & 0.842 & 0.980 & 18.020 \\

GPT-5 (teacher) &
0.931 & 0.913 & 0.985 & 15.590 \\

\rowcolor{gray!20} Top student models &
  &   &   &   \\

llama-3.1 8B &
0.905 & 0.892 & 0.974 & 15.980 \\

llama-3.1 8B + AIQAed labels &
0.914 & 0.901 & 0.976 & 16.146 \\

qwen2.5 7B + AIQAed labels &
0.902 & 0.886 & 0.979 & 16.183 \\

qwen2.5 32B + AIQAed labels &
0.908 & 0.893 & 0.980 & 16.086 \\

\rowcolor{green!15}
llama-3.2 3B + AIQAed labels &
0.915 & 0.901 & 0.977 & 16.350 \\
\end{tabular}

\end{table}

\paragraph{Analysis.}
Several observations emerge from the offline results. First, the GPT-5 \cite{openai2025gpt5} teacher establishes a strong quality ceiling, substantially outperforming the production baseline on precision metrics (P-T@5: 0.931 vs.\ 0.824). Second, AIQA-curated training labels consistently improve student model quality: Llama-3.1 8B improves from 0.905 to 0.914 on P-T@5 when trained with AIQA labels, confirming the value of automated quality assurance in the distillation pipeline. Third, the compact Llama-3.2 3B model with AIQA labels achieves the highest student-model precision (P-T@5: 0.915, P-T@20: 0.901), matching or exceeding larger student variants while offering significant latency and cost advantages at serving time. Finally, we note that all generative policies exhibit lower recall density than the production baseline, reflecting a deliberate trade-off: the generative system produces more targeted, compositional queries that yield smaller but more relevant recall sets.

\subsection{Online Evaluation}
Top-performing policies from offline evaluations are promoted to large-scale A/B experiments against the production baseline, serving hundreds of thousands of users over a product catalog spanning tens of millions of items. Initial online experiments leverage the GPT-5 teacher model for both theme and keyword generation, as infrastructure constraints prevented earlier in-house deployment of the fine-tuned keyword generator. All online policies leverage the DeBERTa cross-encoder for relevance filtering (Section~4.4).

\paragraph{Primary Results.}
We observe statistically significant improvements across key engagement metrics, summarized in Table~\ref{tab:online}.

\begin{table}[t]
\centering
\caption{Online A/B experiment results. Each visit may contain multiple page views.}
\label{tab:online}
\begin{tabular}{lcc}
\toprule
\textbf{Metric} & \textbf{Relative Lift} & \textbf{$p$-value} \\
\midrule
Cart adds per page view & +2.7\% & 0.0001 \\
Cart adds per visit & +1.0\% & 0.01 \\
Alignment with past order behavior & +2.3\% & --- \\
\bottomrule
\end{tabular}
\end{table}

The +2.7\% lift in cart adds per page view ($p = 0.0001$) and +1.0\% lift in cart adds per visit ($p = 0.01$) demonstrate that generative recommendations drive measurable purchasing behavior. 

\paragraph{Long-Tail Personalization.}
A key motivation for the constrained generative approach is improved coverage of long-tail user interests that are available in the catalog taxonomy but underserved by curated content. We observe that the current taxonomy size, paired with RAG, provides good coverage over tail categories and products, enabling the system to surface relevant carousels for niche preferences difficult to address through manual curation alone.

\paragraph{Category-Level Insights.}
Analyzing cart adds per impression at the carousel level, we find that carousels related to fruits, dairy, and snacks are the top-performing categories. This suggests that generative personalization is particularly effective for high-frequency, preference-driven grocery categories where tastes vary widely.

\paragraph{Latency.}
Despite the addition of LLM inference in the recommendation pipeline, we observe no significant change in overall end-to-end latency. This is achieved through aggressive caching of generated themes and keywords, which amortizes the cost of generation across repeated user sessions.

\subsection{Limitations}
 While this framework demonstrates strong results, several limitations remain. First, supervised fine-tuning can introduce popularity biases, causing the student model to over-index on frequent themes and keywords at the expense of niche user interests. Second, highly personalized keywords improve relevance but reduce product recall density, risking carousel collapse when catalog coverage is insufficient. Third, the reliance on a fixed keyword taxonomy limits the system's ability to surface novel concepts outside its scope. Finally, the cascaded architecture can introduce inter-phase error propagation, where low-quality theme generation in Phase 1 can degrade downstream retrieval. Addressing these limitations through reinforcement fine-tuning \cite{ouyang2022instructgpt}, dynamic taxonomy expansion, and cascaded error correction are active areas of future work.

\section{Conclusion and Future Work}
We introduce a cascaded generative merchandising framework for personalizing e-commerce storefronts. Ordered themes first personalize page structure, with keywords generated per placement theme to power product retrieval. This approach delivers rich personalization while maintaining compatibility with downstream retrieval and ranking systems. We observe meaningful improvement in content quality and user relevance across offline and online evaluations.

Future work will include multi-objective optimization across business goals (ex: balancing relevance with novel discovery), surface expansion beyond storefront, and deeper personalization via real-time and sparse user signals. Additionally, reinforcement fine-tuning (RFT) will enable system self-improvement with closed-loop feedback. Over the long run, we project that traditional ranking models can be directly fused as rewards within RFT post-training, bringing recommendations fully into the generative paradigm.

\begin{acks}
We would like to extend deep thanks to our cross-functional partners Logan Murdock, Dhruv Khanna, Jingying Zhou, Roy Li, Aref Kashani, Shayaan Nadeem, Amish Popli, Lauren Downey, Shrikar Archak, Brett Brownell, and Max Silberstein, who provided critical design input, accelerated our evaluation framework, and unlocked rapid online experimentation. Additional thanks to Pramod Adiddam for steady leadership and support that enabled the team to push this work forward. 
\end{acks}

\bibliographystyle{ACM-Reference-Format}
\bibliography{references}

\end{document}